\documentclass[10pt,twocolumn,letterpaper]{article}

\usepackage{iccv}              

\definecolor{iccvblue}{rgb}{0.21,0.49,0.74}
\usepackage[pagebackref,breaklinks,colorlinks,allcolors=iccvblue]{hyperref}
\usepackage{makecell}
\usepackage{multirow}

\usepackage{booktabs}
\usepackage{amssymb}
\usepackage[misc]{ifsym} 
\newcommand{\myparagraph}[1]{{\vspace{0.6mm}\noindent\bf #1}}
\def\paperID{8071} 
\def\confName{ICCV}
\def\confYear{2025}
\title{CharaConsist: Fine-Grained Consistent Character Generation}

\author{Mengyu Wang$^{1,3}$, Henghui Ding$^2$\footnotemark[2] , Jianing Peng$^{1,3}$, Yao Zhao$^{1,3}$, Yunpeng Chen$^4$, Yunchao Wei$^{1,3}$\footnotemark[2] \\
$^1$Institute of Information Science, Beijing Jiaotong University, Beijing, China\\
$^2$Institute of Big Data, Fudan University, Shanghai, China\\
$^3$Visual Intelligence + X International Joint Laboratory of the Ministry of Education\\
$^4$Alkaid Pte. Ltd., Singapore\\
{\tt\small mengyu.wang@bjtu.edu.cn, henghui.ding@gmail.com, wychao1987@gmail.com}
}

\begin{document}

\twocolumn[{%
\renewcommand\twocolumn[1][]{#1}%
\maketitle
\begin{center}
    \centering
    \captionsetup{type=figure}
    \vspace{-5mm}
    \includegraphics[width=1.0\linewidth]{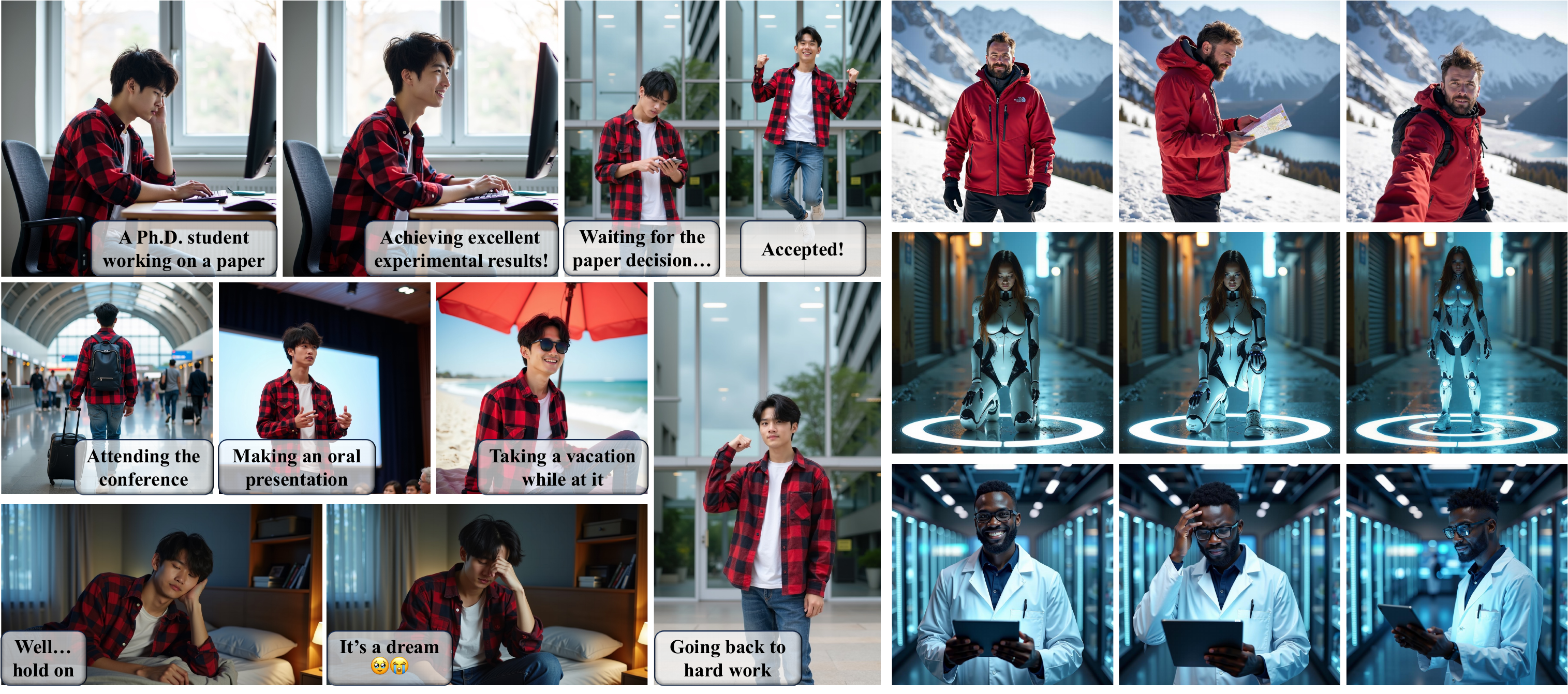}
    \vspace{-7mm}
    \captionof{figure}{\textbf{\textit{CharaConsist} achieves fine-grained consistency maintaining.} The left part shows that, \textit{CharaConsist} enables more flexible storytelling, supports controllable scene transitions, and achieves character consistency within a fixed scene, across different scenes, and across resolutions. The right part highlights \textit{CharaConsist}'s ability to preserve extensive background details with complete consistency.}
    \label{fig:teaser}
    \vspace{2mm}
\end{center}%
}]
\renewcommand{\thefootnote}{\fnsymbol{footnote}}
\footnotetext[2]{Corresponding authors: Henghui Ding and Yunchao Wei.}

\begin{abstract}
In text-to-image generation, producing a series of consistent contents that preserve the same identity is highly valuable for real-world applications. 
Although a few works have explored training-free methods to enhance the consistency of generated subjects, we observe that they suffer from the following problems.
First, they fail to maintain consistent background details, which limits their applicability. Furthermore, when the foreground character undergoes large motion variations, inconsistencies in identity and clothing details become evident. To address these problems, we propose CharaConsist, which employs point-tracking attention and adaptive token merge along with decoupled control of the foreground and background.
CharaConsist enables fine-grained consistency for both foreground and background, supporting the generation of one character in continuous shots within a fixed scene or in discrete shots across different scenes.
Moreover, CharaConsist is the first consistent generation method tailored for text-to-image DiT model. Its ability to maintain fine-grained consistency, combined with the larger capacity of latest base model, enables it to produce high-quality visual outputs, broadening its applicability to a wider range of real-world scenarios.
The source code has been released at
\href{https://github.com/Murray-Wang/CharaConsist}
{CharaConsist.git}.
\end{abstract}   
\vspace{-6mm}\section{Introduction}
\label{sec:intro}

\begin{figure*}
  \centering
  \includegraphics[width=1\linewidth]{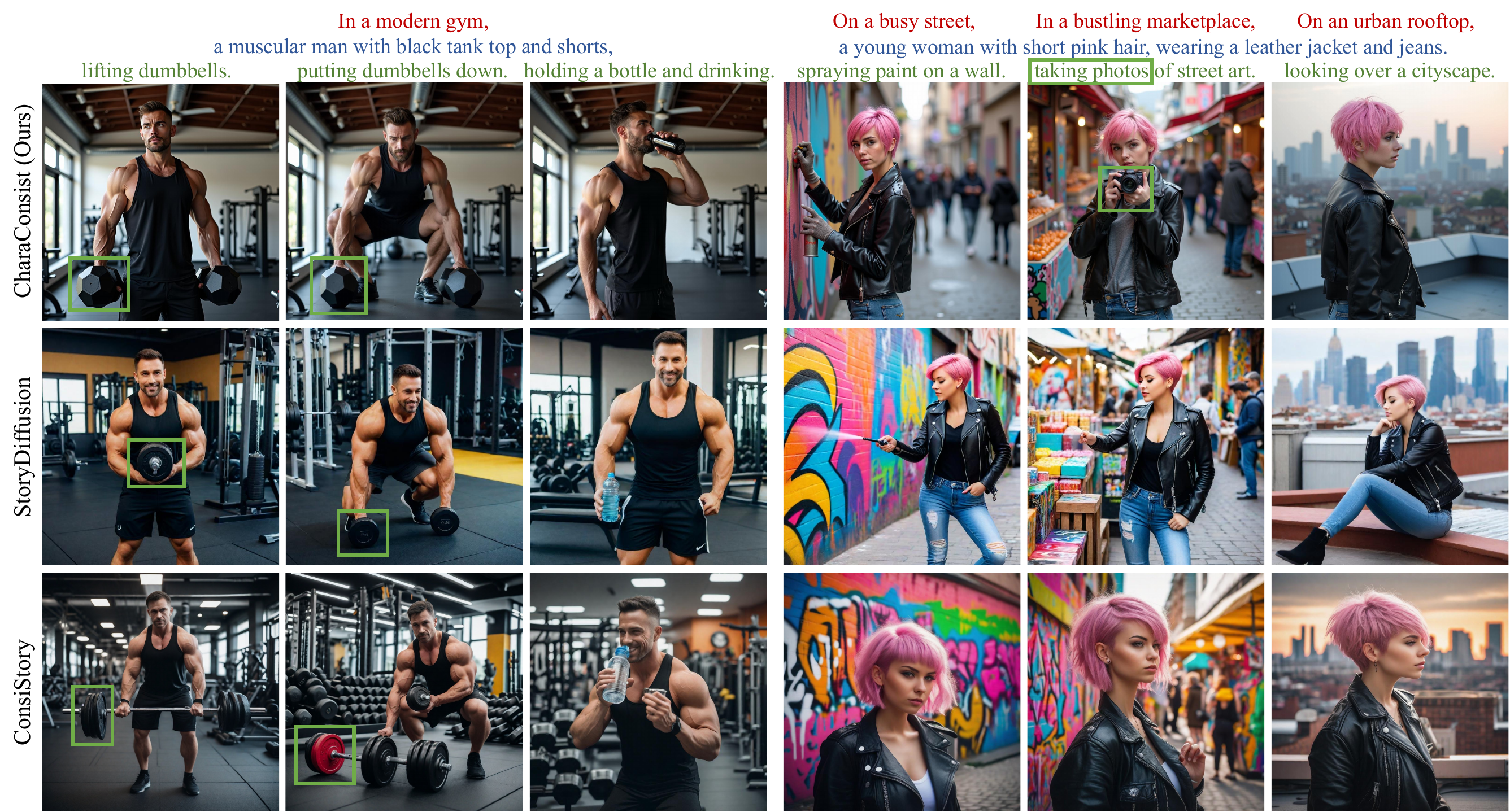}
  \vspace{-6.6mm}
  \caption{\textbf{Qualitative comparison with StoryDiffusion}~\cite{storydiffusion} \textbf{and ConsiStory}~\cite{consistory}. For continuous shots within a fixed scene (the left part), previous methods cannot maintain consistent background and foreground details. For discrete shots across different scenes (the right part), previous methods either exhibit a copy-paste effect or suffer from inconsistency for the characters. In contrast, our \textit{CharaConsist} maintains fine-grained consistency for both characters and environments, while also providing greater flexibility in character actions.}
  \label{fig:compare}
  \vspace{-2.5mm}
\end{figure*}

Text-to-image (T2I) generation has advanced significantly in recent years \cite{ldm,sdxl}, especially with the widespread emergence of Diffusion Transformers (DiTs) \cite{dit,sd3,flux}.
Their successful scaling-up has led to significant improvements in both image quality and text alignment, unlocking new possibilities for many real-world applications. A key challenge is generating multiple images while ensuring subject consistency, a crucial requirement that has gained increasing attention~\cite{storydiffusion,consistory,1p1s}. Consistency is essential for applications such as visual storytelling, virtual character design, and long video generation, where continuity across images enhances usability and realism.

Considering computational resources and time consumption, recent studies have explored training-free methods \cite{storydiffusion,consistory,1p1s} to achieve consistent T2I generation. These methods introduce inter-image information sharing during generating a batch of images to enhance consistency and focus primarily on achieving the visual effect of \textbf{one character transitioning across different backgrounds}.
However, this task formulation remains overly coarse, whereas many real-world applications demand \textbf{finer-grained consistency}.
For example, as shown in \cref{fig:teaser},
visual storytelling often requires generating a sequence of coherent story shots within the same scene. 
Similarly, in video-related applications, ensuring fully consistent keyframes is critical for guiding long video synthesis. Failure to do so can lead to flickering artifacts, compromising the final output quality.

Existing methods~\cite{storydiffusion,consistory,1p1s} face significant challenges when addressing such fine-grained requirements.~First, their results exhibit \textbf{inconsistency of the static background}. When generating a series of continuous shots, \eg, the gym scene in \cref{fig:compare}, these methods failed to maintain consistency of neither the background instruments, nor the foreground dumbbell, which could lead to flickering artifacts when the generated results serving as video keyframes.
Second, they suffer from a \textbf{trade-off between foreground consistency and action variation}. As shown in the right part of \cref{fig:compare}, the generated characters either exhibit a nearly copy-paste effect, deviating from the unique action description of each image, or suffer from inconsistency in details like clothing and hairstyle.~Furthermore, these methods are inherently constrained by the \textbf{underpowered base model}.
Most existing approaches are built on SDXL \cite{sdxl}, an earlier T2I model that lags behind state-of-the-art Diffusion Transformers (DiTs)~\cite{dit,sd3,flux} in generating high-quality characters and complex scenes. Additionally, their UNet-specific implementation strategies cannot be directly adapted to newer DiT architectures.

To address these challenges, we propose \textit{{CharaConsist}}, a training-free method built on FLUX.1~\cite{flux} to achieve the fine-grained consistency.
First, we introduce point-tracking attention. This module first establishes positional correspondences across images and subsequently re-encode the positional embeddings during the inter-image information sharing process. Point-tracking attention enables the automatic tracking of critical features across images, therefore not only enhancing consistency but also providing greater flexibility in character actions.
Second, while re-encoding the positional embeddings mitigates the impact of positional shifts, it may inadvertently disrupt the original local geometric structure among tokens. To address this, we further introduce an adaptive token merge strategy, which refines the attention outputs by dynamically aggregating tokens based on their similarity and positional correspondences established in the first step.
Third, to accommodate varying requirements for background consistency across different application scenarios, we introduce a foreground-background mask, which is derived from the attention differences between image tokens and text tokens corresponding to foreground and background elements. By leveraging this distinction, our approach enables decoupled control over foreground and background consistency, allowing for more flexible adaptation to diverse generation needs.

In experiments, we compare the proposed \textit{CharaConsist} with previous training-free consistent T2I generation methods, including ConsiStory \cite{consistory} and StoryDiffusion \cite{storydiffusion}, as well as training-dependent identity-reference generation methods, such as IP-Adapter \cite{ip-adapter} and PhotoMaker \cite{photo-maker}. The results demonstrate that \textit{CharaConsist} significantly outperforms these methods in both foreground and background consistency, underscoring its effectiveness in fine-grained consistent generation.
In summary, our key contributions are as follows:
\begin{itemize}
    \item We highlight that existing consistent generation works adopt relatively coarse task formulations, primarily concentrating on transitions of similar characters across different scenes, while lacking the fine-grained consistency in character details and background environments.
    \item We propose \textit{CharaConsist}, which leverages point tracking and mask extraction to automatically perceive critical features, thus achieving fine-grained consistency, making it well-suited for real-world applications.
    \item \textit{CharaConsist} is the first training-free consistent T2I generation method built on a DiT model. It operates without additional training or extra modules, enabling the efficient utilization of large DiT models with billions of parameters. Moreover, the proposed DiT-based point tracking and mask extraction techniques can serve as effective tools for related tasks such as image editing.
\end{itemize}

\section{Related Work}
\label{sec:related}
\subsection{Diffusion-based T2I Generation}
With the advent of DDPM \cite{ddpm}, diffusion models \cite{diffusion2015} have gradually surpassed GANs \cite{gan} to become the dominant technology in visual generation.
Earlier diffusion models were primarily based on the UNet \cite{unet} architecture, such as SD 1.x \cite{ldm}, SD 2.x, and SDXL \cite{sdxl}.
These models process visual information through convolutional \cite{resnet} and self-attention \cite{transformer} blocks, and capture textual prompt information through cross-attention with text tokens encoded by CLIP \cite{clip}, enabling text-to-image generation.

More recently, Diffusion Transformers (DiTs) \cite{dit} were introduced, where the authors replaced the commonly used UNet with a Transformer \cite{transformer} and demonstrated its superior scalability.
Over the past one year, numerous DiT-based T2I \cite{hunyuan,sd3,flux,shuai2024survey} and T2V \cite{cogvideox,hunyuan-video,AnyI2V,shuai2025free} models have emerged and demonstrated a superior ability on producing visually stunning results.
Among them, the latest T2I DiTs, such as SD3.x \cite{sd3} and FLUX.1 \cite{flux}, adopt similar multimodal-transformer architectures. These models use the T5 encoder \cite{t5} to encode text tokens and a vae \cite{vae,vqgan} to encode image tokens, and then concatenate these tokens into a single sequence to perform attention.
Compared to UNet-based T2I models, these models no longer feature separate self-attention and cross-attention, making it challenging to directly adapt previous methods to DiTs, which often manipulate the specific layers in the UNet.

\vspace{-.5mm}
\subsection{Training-free Consistent Generation}
\vspace{-1mm}

As a relatively new task, related research is primarily limited to three works: 1P1S \cite{1p1s}, StoryDiffusion \cite{storydiffusion}, and ConsiStory \cite{consistory}.
1P1S is built on the consistency from text tokens, which concatenates the textual prompts of multiple images and encodes them together.
However, this approach is limited by the maximum sequence length of the text encoder.
As a result, it's challenging to employ it to scenarios requiring relatively long text prompts, such as the diverse characters and scenes generation in this work.

StoryDiffusion and ConsiStory share similar idea, introducing inter-image attention as core technique to enhance consistency.
They randomly sample a subset of tokens from a batch of images and use them as shared keys and values in the attention operation for each image. 
However, we observe that such mechanism has low information retrieval efficiency. This results in two major issues: first, they require at least 2-4 images to be generated in parallel to enhance inter-image associations, leading to significant additional memory consumption. Second, the consistency of their results remains unsatisfactory.
Therefore, our proposed \textit{CharaConsist} introduces point-tracking-based attention and token merging, which can adaptively retrieves critical information to enhance consistency, without the need for any image parallelism.

Additionally, similar to our adaptive token merge module, ConsiStory also introduces similarity-based feature injection.
However, their similarity matching relies on DIFT \cite{dift}, a technique based on specific layers and steps of the UNet architecture. However, we found that this approach does not perform effectively on the DiT model.

\vspace{-1mm}
\subsection{Identity-reference Generation}
\vspace{-1mm}
Similar to consistent generation, identity-reference generation aims to align the identity of the generated results with the input reference face image.
These works follow two main approaches: The first one involves introducing pre-trained image encoders and training additional projection module to embed the identity information into the generation process, such as IP-Adapter \cite{ip-adapter} and PhotoMaker \cite{photo-maker}.
The second approach, widely known as personalization, depends on per-subject training. These methods encodes a specific subject's identity into a new textual embedding \cite{textual-inversion} or directly into the model's parameters \cite{dreambooth,custom-diff}, ultimately making the generated results overfit to a fixed identity.

However, these methods focus solely on facial identity preserving and cannot ensure consistency in other attributes such as clothing or scenes.
Additionally, they require training on large-scale collected datasets, or performing per-subject training, which imposes high demands on computational resources or time consumption.
\section{Method}
\label{sec:method}
\begin{figure*}
  \centering
  \includegraphics[width=1\linewidth]{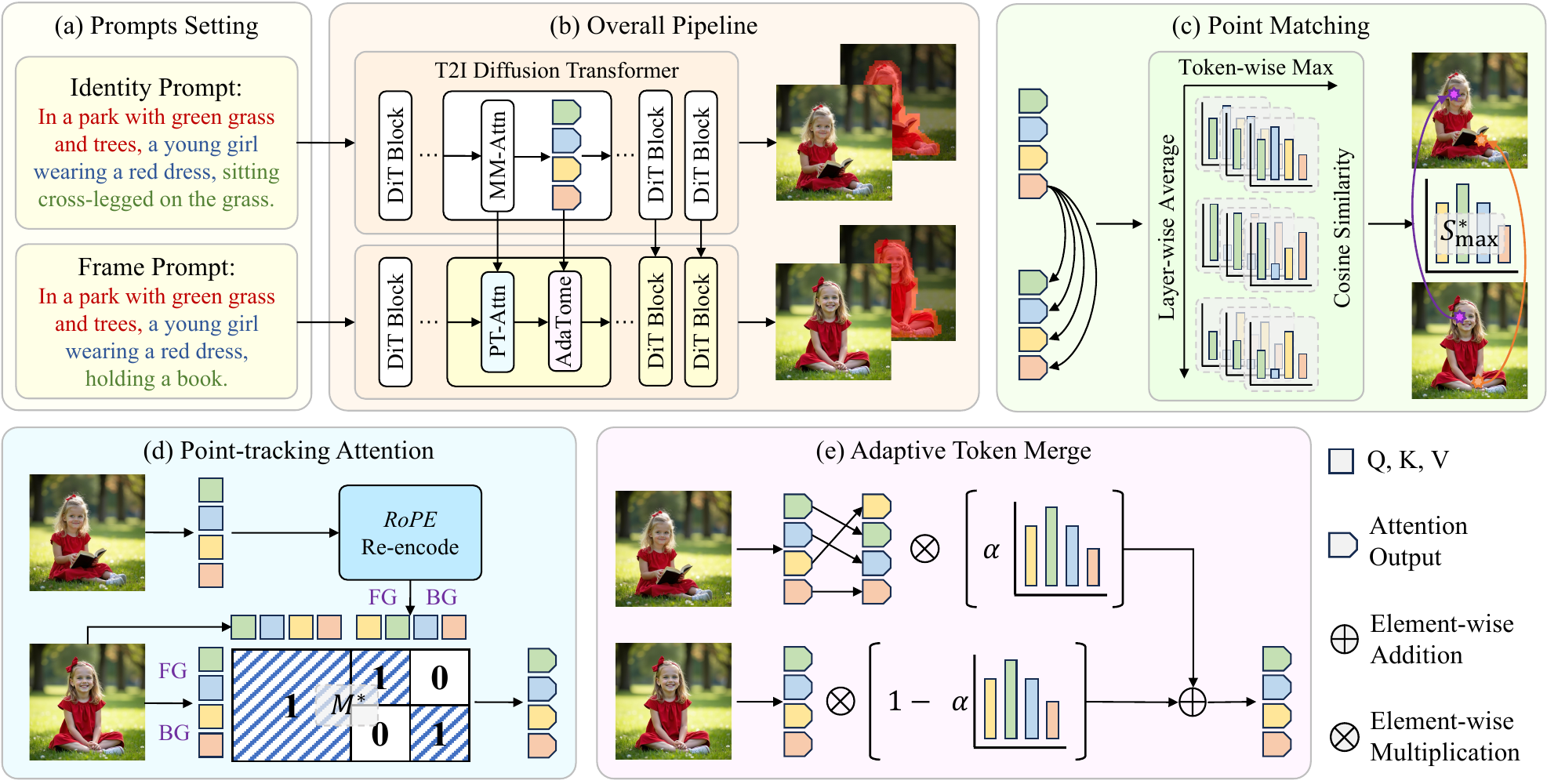}
  \vspace{-6.6mm}
  \caption{\textbf{An overview of our \textit{CharaConsist}}. (a) shows an example of the textual prompts. (b) is the overall pipeline of our method, in which we replace the original multimodal attention with our proposed point-tracking attention, and further introduce an adaptive token merge module, as detailed in (d) and (e). (c) shows the point matching strategy of our method.}
  \label{fig:method}
  \vspace{-3mm}
\end{figure*}

In this section, we first provide a brief review and analysis of the core technique, inter-image attention, used to enhance consistency in previous works, along with some technical details of the base model FLUX.1 in \cref{sec:preliminary}.
Then, we introduce our proposed \textit{CharaConsist} in \cref{sec:characonsist}. \Cref{fig:method} presents the schematic of the overall framework and individual components of our method.

\subsection{Preliminary}
\label{sec:preliminary}

\myparagraph{Inter-image Attention}
In previous works, inter-image attention is a core technique used to enhance consistency. During the generation of a batch of images, inter-image attention works by concatenating keys and values from neighboring images into each image’s self-attention process, enabling information sharing across images. This process helps maintain coherence in the generated sequence by promoting cross-image relationships.

\begin{figure}[t]
  \centering
  \includegraphics[width=1\linewidth]{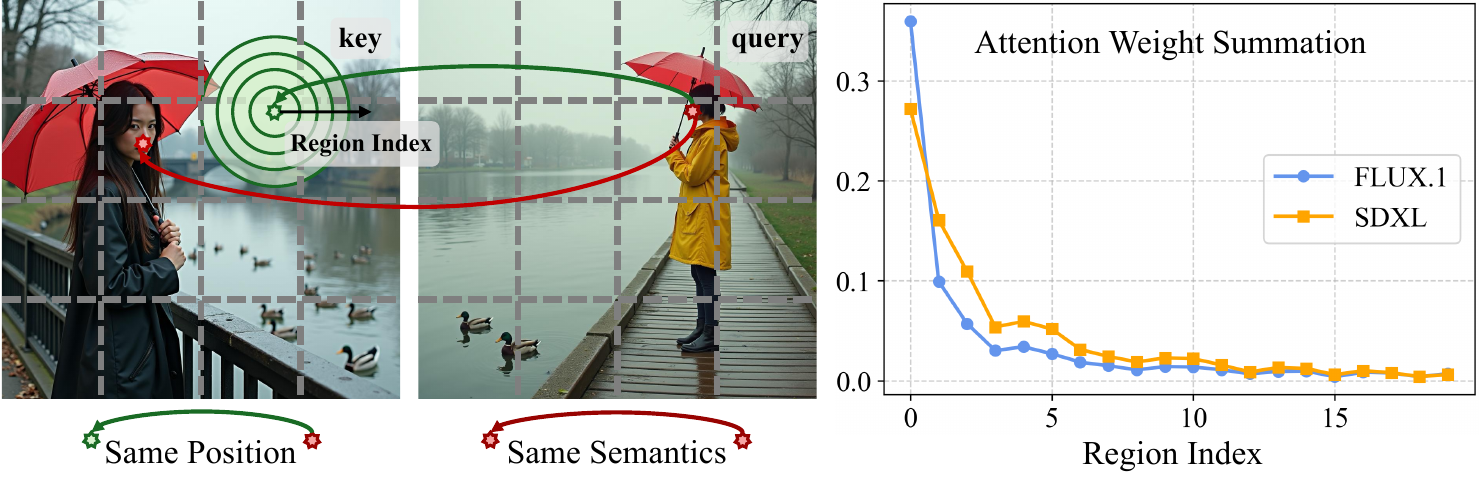}
  \vspace{-7.6mm}
  \caption{\textbf{The locality bias in inter-image attention}. Both SDXL \cite{sdxl} and FLUX.1 \cite{flux} tend to allocate more attention weights to spatially nearby regions instead of more semantically related ones.}
  \label{fig:local_attn}
  \vspace{-5mm}
\end{figure}

However, we observe that although attention module has a global receptive field, directly applying inter-image attention struggles to capture the necessary correspondences for maintaining consistency.~To intuitively demonstrate this, as shown in \cref{fig:local_attn}, we generate images with similar content but varying layouts and compute their averaged inter-image attention weights using both the UNet-based SDXL~\cite{sdxl} and the DiT-based FLUX.1~\cite{flux}. Using the query coordinate as the origin, we divide each image into concentric rings of equal area and calculate the attention weight summation within each ring. The results show that as the distance from the query coordinate increases, the attention weight summation of the corresponding region decreases significantly. This phenomenon indicates that the model exhibits a locality bias, tending to allocate more attention weights to spatially nearby regions rather than to more semantically related ones. As a result, the model struggles to maintain consistency across varying layouts, failing to capture distant but contextually important features.

The inconsistency observed in previous methods, as shown in \cref{fig:compare}, could largely stem from this locality bias. When characters in different images undergo large motion variations or occupy different positions within the image, their relative distance could increase significantly. As a result, the inter-image attention yields lower response intensity for these distant regions, making it challenging to gather sufficient information to ensure consistency. Building on this observation, our method begins with point tracking and incorporates several techniques to mitigate the effects of the positional shift, ensuring more robust and consistent representations across varying image layouts.

\setlength{\abovedisplayskip}{4pt}
\setlength{\belowdisplayskip}{4pt}

\myparagraph{FLUX.1}
FLUX.1~\cite{flux} is a latent-space DiT-based T2I model with a multimodal transformer as its core architecture. It includes two types of transformer blocks: the double block and the single block, with both blocks taking text and image tokens as input.

Although the two blocks differ in structure and parameter allocation, their core mechanism remains the same: the concatenated text and image tokens undergo global attention.
As a training-free method, our primary focus is on the attention process within the model.
Therefore, we omit the structural differences between the two types of blocks and formulate the attention process in both as a unified form:
\begin{align}
    X_{mm} &= \text{cat}[X_{t}, X_{i}] \in \mathbb{R}^{(l+hw) \times d}, X \in \{Q, K, V\}, \\
    W_{mm} &= \frac{Q_{mm} \cdot K_{mm}^T}{\sqrt{d}} \in \mathbb{R}^{(l+hw) \times (l+hw)},\\
    H_{mm} &= \text{Softmax}(W_{mm}) \cdot V_{mm} = \text{cat}[H_{t}, H_{i}], \label{equ:attn_out}
\end{align}
where $\text{cat}[\cdot]$ refers to concatenation. The subscripts $mm$, $t$ and $i$ stand for ``multimodal'', ``text'', and ``image'', respectively. $l$ is the length of text tokens. $d$ is the dimension of each attention head. $(h, w)$ represents the resolution of the image tokens, which is $1/16$ of the generated images.
The number of attention head is omitted for simplicity.

\subsection{CharaConsist}
\label{sec:characonsist}
\myparagraph{Pipeline Definition}
We divide the consistent generation pipeline into two parts: identity image generation and frame images generation. The identity image generation follows the original generation process without modifications, and stores intermediate variables such as keys, values, and attention outputs from specific layers at certain timesteps.
The frame images then achieve consistency with the identity image by accessing these stored variables.

Unlike previous methods that require generating at least 2 to 4 batched images in parallel to obtain identity information, \textit{CharaConsist} needs only a single identity image. This advantage largely arises from the tracking mechanism in our method, which effectively uses the identity information from a single image, eliminating the need for parallel processing for augmentation.
As a result, this approach significantly reduces the additional GPU memory overhead.

\myparagraph{Point Matching}
The goal of point matching is to identify, for each point in the frame image, the semantically corresponding point within the identity image, thereby guiding subsequent attention and token merging processes.
For example, the facial features of the character in the frame image should align with those in the identity image.
Previous work DIFT~\cite{dift} demonstrates that semantic correspondence can be determined by measuring the similarity of intermediate features from the UNet in diffusion models at a specific timestep.
However, we observe that this approach failed in FLUX.1.
Specifically, we test the cosine similarity of outputs from both intermediate transformer blocks and attention layers in FLUX.1.
For the transformer block outputs, we found that the matching points remain fixed at the same position, failing to capture semantic correspondence.
In contrast, for attention layer outputs, the matching points exhibit significant fluctuations across layers, timesteps, and samples, making stable and accurate matching difficult.
Furthermore, we observe that while matching relationships across layers vary significantly, the similarity distributions exhibit consistent characteristics. Based on this observation, we average the similarity scores across layers at the same timestep, formulated as:
\begin{align}
    &S^* = \frac{1}{N} \sum_n\frac{H_{i,frm,n} \cdot H_{i,id,n}^T}{\|H_{i,frm,n}\|_2 \cdot \|H_{i,id,n}\|_2}, \\
    &map^*(\cdot), S^*_{\text{max}} = \text{argmax}(S^*), \text{max}(S^*), \label{equ:sim_max}
\end{align}
where $N$ and $n$ are the number and index of layers, respectively. $map^*(\cdot)$ is the matching results and $map^*(j)=k$ means the $j$-th point in frame image is matched with the $k$-th point in identity image.
$S^*_{\text{max}}$ is the averaged cosine similarity of the matched token pair.
This approach provides stable semantic similarity results and accurate point matching relationships.
Results and further details can be found in \cref{fig:point_mask} and the supplementary materials.

\myparagraph{Mask Extraction}
Our mask extraction strategy is built on comparing the attention weight of image tokens to foreground text tokens against the weight to background text tokens.
Specifically, we format the textual prompt as the background description first, followed by the foreground description, as the example shown in \cref{fig:method}. We then obtain their respective token lengths, denoted as $l_{bg}$ for background and $l_{fg}$ for foreground.
Next, we extract the segment of attention weights where image tokens serve as queries and text tokens as keys, represented as:
\begin{align}
    A_{i2t} &= \text{Softmax}(W_{mm}[l:, :l_{bg}+l_{fg}]).
\end{align}
\indent We apply the Softmax operation only within the actual length of the text sequence to prevent the impact of the image keys and the special text key ``\texttt{<}s\texttt{>}''.
Here, we observe a similar issue to the point matching process: the attention weights corresponding to each word exhibit significant fluctuations across layers. To address this, we average the attention weights within the foreground and background text sequence separately, and also across different layers. By comparing these averaged attention weights, we derive the robust foreground mask, represented as:
\begin{align}
    A^*_{i2bg} &= \frac{1}{N \times l_{bg}}\sum_n \sum_{j=0}^{l_{bg}-1} A_{i2t,n}[:, j],\\
    A^*_{i2fg} &= \frac{1}{N \times l_{fg}}\sum_n \sum_{j=l_{bg}}^{l_{bg}+l_{fg}-1} A_{i2t,n}[:, j],\\
    M &= A^*_{i2fg} > A^*_{i2bg} \in \{0,1\}^{hw}.
\end{align}

\myparagraph{Point-tracking Attention}
After obtaining the point-matching relationships and the foreground masks, we introduce point-tracking attention. Specifically, we first extract the foreground token indices in the frame image, along with the matched token indices in the identity image:
\begin{align}
    C_{fg,frm} &= \text{nonzero}(M_{frm}), \\
    C_{fg,id} &= map^*(C_{frm}),
\end{align}
where $\text{nonzero}(\cdot)$ refers to get the indices of non-zero elements.
Based on the indices, we extract the keys and values from the identity image and re-encode the positional embedding of the keys as follows:
\begin{align}
    &X_{fg,id} = X_{i,id}[C_{fg,id}], X \in \{K^0, V\}, \\
    &K'_{fg,id} = RoPE(K^0_{fg,id}, C_{fg,frm}),
\end{align}
where $K^0$ represents the keys without positional embedding, $RoPE$ indicates the Rotary Position Embedding \cite{rope} used in FLUX.1.
We further concatenate the original keys of frame image, the re-encoded foreground keys of identity image, and the background keys of identity image to get the keys of our point-tracking attention, formulated as:
\begin{align}
    K^*_{frm} &= \text{cat}[K_{mm,frm}, K'_{fg,id},K_{bg,id}],\\
    V^*_{frm} &= \text{cat}[V_{mm,frm}, V_{fg,id},V_{bg,id}].
\end{align}
\indent The operations related to background are optional, depending on whether users want to maintain the background unchanged. Then the attention process is represented as:
\begin{align}
    &W^*_{frm} = \frac{Q_{mm,frm} \cdot K^{*T}_{frm}}{\sqrt{d}} + log(M^*),\\
    &H^*_{mm,frm} = \text{Softmax}(W^*_{frm}) \cdot V^*_{frm},
\end{align}
where $M^*$ is the attention mask, making the foreground tokens in frame image only attend to the foreground tokens in identity image, and vice versa.
This approach effectively leverages the model's characteristics, as $RoPE$ is applied directly at each attention layer, allowing to store the keys without positional embedding and encode it later as wished.

\myparagraph{Adaptive Token Merge}
Although the re-encoding of the positional embedding enhances consistency through attention, it may also cause disruptions in the locally geometric relationships of tokens due to positional rearrangement. Additionally, a few tokens may not be matched perfectly, leading to information loss.
Considering that attention output is a weighted sum of all values and carries more global information, it helps to compensate for such information loss.
Therefore, we introduce adaptive token merge, where the attention outputs of the frame image are interpolated with those from the identity image.

In this process, we first extract the segment attention output corresponding to the foreground image tokens, from both the identity image and frame image. 
We then rearrange the attention output of identity image based on the point-matching relationships and perform interpolation:
\begin{align}
    &H_{fg,id}\ \ \ \ \! = H_{i,id}[C_{fg,id}],\\
    &H^*_{fg,frm} = H^*_{i,frm}[C_{fg,frm}],\\
    &H^{**}_{fg,frm} = (1 - \alpha S^*_{\text{max}}) H^{*}_{fg,frm} + \alpha S^*_{\text{max}}H'_{fg,id},
\end{align}
where $\alpha\in[0,1]$ is a hyper-parameter and decays with the timestep in the generation process.
This interpolation process incorporates the averaged similarity $S^*_{\text{max}}$ obtained from \cref{equ:sim_max} as weights, effectively suppressing tokens with low matching confidence and preventing negative impact.

\begin{figure*}
  \centering
  \includegraphics[width=1\linewidth]{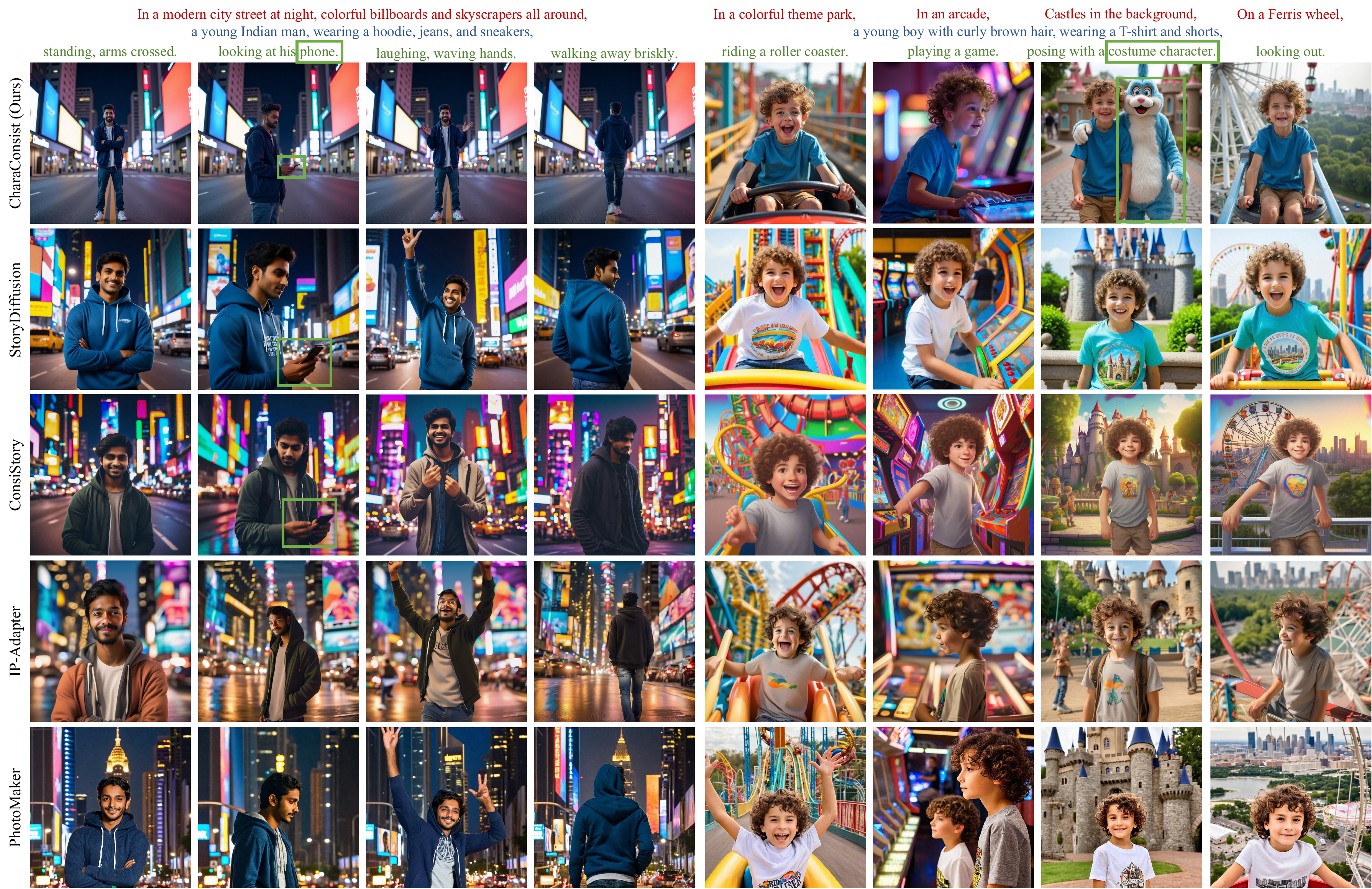}
  \vspace{-6.6mm}
  \caption{\textbf{Qualitative comparisons with previous consistent generation methods and identity-reference methods.} Our method demonstrates superior performance in maintain consistency of multiple aspects such as character identity, clothing and background scenes.}
  \label{fig:quali}
  \vspace{-3mm}
\end{figure*}

\section{Experiments}
\label{sec:exp}

\subsection{Experimental Setups}
\myparagraph{Evaluation Datasets}
This work focuses on consistent generation for characters and environments, while existing benchmarks lack features suited for such tasks. Therefore, we used GPT-4 to generate a series of T2I prompts tailored for such application scenario.
Specifically, we instructed GPT-4 to create multiple groups of prompts in the format: ``[Environment], [Character], [Action].'', with contents vary across groups. 
We proposed two evaluation tasks: \textbf{background maintaining} and \textbf{background switching}. In the former, both ``[Environment]'' and ``[Character]'' remain consistent within a group of prompts, while ``[Action]'' varies. In the latter, only the ``[Character]'' remains unchanged.
Every group consists of 5-8 prompts.
For each of the two-types tasks, we generated over 200 T2I prompts.

\begin{table*}[t]
\small
\setlength{\tabcolsep}{4.5pt}
\centering
  \begin{tabular}{lccccccccccc}
    \Xhline{0.8pt}
    \multirow{2}{*}{Method} & \multicolumn{5}{c}{Background Maintaining} & \multicolumn{4}{c}{Background Switching} & \multicolumn{2}{c}{Image Quality} \\
    & CLIP-T & CLIP-I & CLIP-I-fg & CLIP-I-bg & ID Sim & CLIP-T & CLIP-I & CLIP-I-fg & ID Sim & IAS & IQS \\
    \Xhline{0.5pt}
    IP-Adapter~\cite{ip-adapter} & 0.358 & 0.891 & \textbf{0.883} & 0.881 & 0.664 & 0.357 & 0.833 & 0.878 & 0.590 & 0.811 & 0.981\\
    PhotoMaker~\cite{photo-maker} & 0.365 & 0.891 & 0.881 & 0.886 & \textbf{0.730} & \textbf{0.371} & 0.824 & 0.876 & \textbf{0.674} & 0.705 & 0.972 \\
    \Xhline{0.5pt}
    StoryDiffusion~\cite{storydiffusion} & \underline{\textbf{0.370}} & 0.902 & 0.882 & 0.895 & 0.642 & \underline{0.367} & \underline{\textbf{0.853}} & 0.873 & 0.563 & 0.777 & 0.978 \\
    ConsiStory~\cite{consistory} & 0.368 & 0.903 & 0.876 & 0.895 & 0.578 & 0.365 & 0.846 & 0.868 & 0.502 & 0.815 & 0.978 \\
    \textbf{\textit{CharaConsist}} (ours) & 0.356 & \underline{\textbf{0.910}} & \underline{\textbf{0.883}} & \underline{\textbf{0.916}} & \underline{0.647} & 0.360 & 0.840 & \underline{\textbf{0.881}} & \underline{0.587} & \underline{\textbf{0.831}} & \underline{\textbf{0.985}} \\
    
    \Xhline{0.8pt}
  \end{tabular}
  \vspace{-3mm}
  \caption{\textbf{Quantitative comparisons with previous consistent generation methods and identity-reference methods.} The best results within the training-free consistent generation methods are \underline{underlined}, and the best results across all methods are highlighted in \textbf{bold}.}
  \label{tab:quan}
  \vspace{-4mm}
\end{table*}

\myparagraph{Comparison Methods}
Comparison methods include both the training-free consistent generation methods StoryDiffusion \cite{storydiffusion}, ConsiStory \cite{consistory} and the training-dependent identity-reference methods IP-Adapter \cite{ip-adapter}, PhotoMaker \cite{photo-maker}.
The above generated prompts are shared for all methods.
For the identity-reference methods, we first generated the character headshots using the ``[Character]'' part of prompts, and then cropped the face region using RetinaFace \cite{retinaface} as the input reference face image.

\myparagraph{Evaluation Metrics}
Following previous works \cite{consistory,storydiffusion}, we introduce the CLIP text-image similarity (\textbf{CLIP-T}) to measure the alignment between images and textual prompts, along with the pairwise CLIP image similarity (\textbf{CLIP-I}) to measure the full-image consistency.
To better evaluate the characteristics of this work, we introduced three additional metrics.
First, we introduce the decoupled CLIP-I, termed as \textbf{CLIP-I-fg} and \textbf{CLIP-I-bg}, to evaluate the consistency of foreground and background separately.
We employ SAM \cite{sam} to segment the foreground in each generated image, and compute CLIP similarity separately for the masked foreground and masked background images.
We also introduce the identity similarity (\textbf{ID Sim}) to evaluate the identity consistency by compute the facial embedding similarity using RetinaFace \cite{retinaface} and FaceNet \cite{facenet}.
Furthermore, we introduce the image quality score (\textbf{IQS}) and image aesthetics score (\textbf{IAS}) to evaluate the quality of generated images using Q-Align \cite{Q-ALIGN}.

\subsection{Experimental Results}

\begin{figure}[t]
  \centering
  \includegraphics[width=1\linewidth]{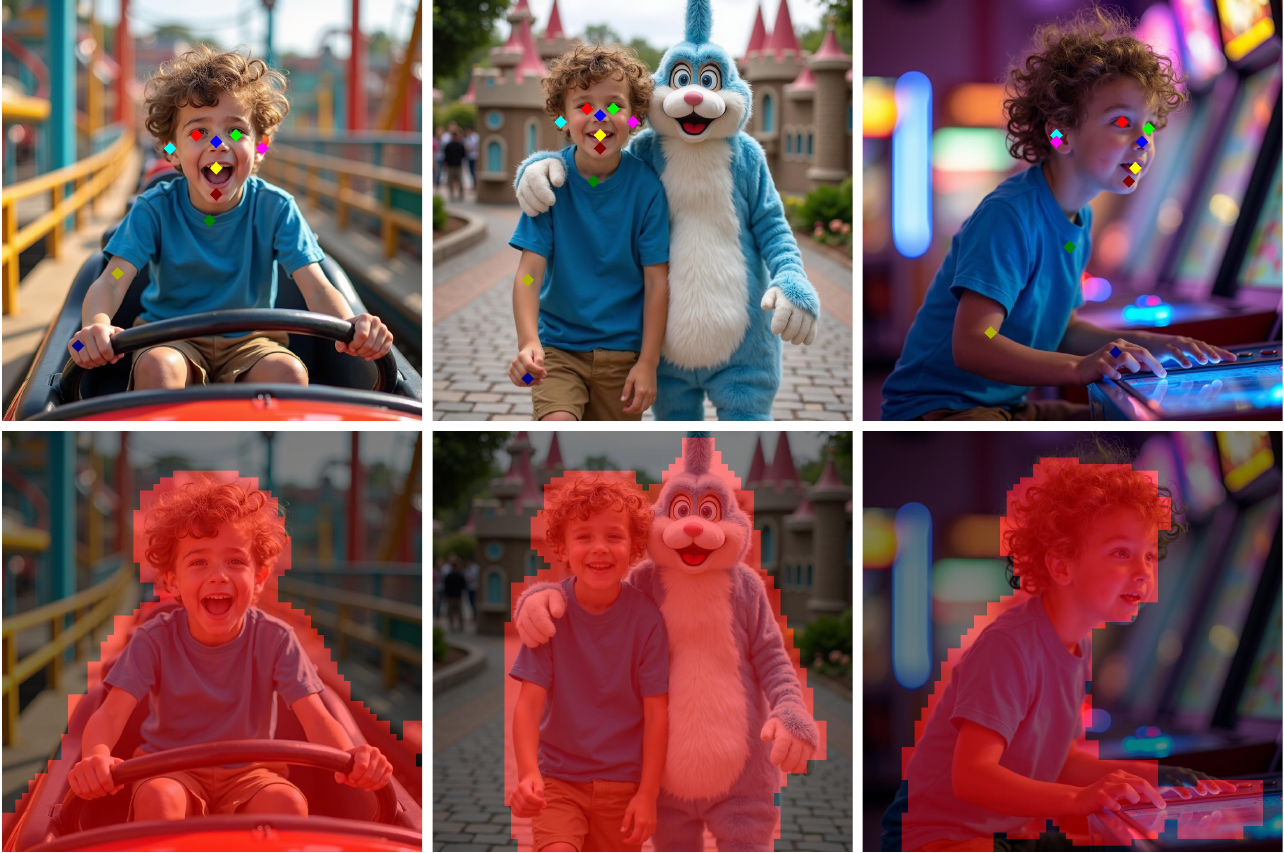}
  \vspace{-7.6mm}
  \caption{\textbf{Point matching and mask extraction results.} The points in the first image are manually selected, while those in the subsequent images are determined by the matching relationship.}
  \label{fig:point_mask}
  \vspace{-3.6mm}
\end{figure}

\myparagraph{Qualitative Results}
In \cref{fig:quali}, we present the results of our \textit{CharaConsist} and all comparison methods on both background maintaining and background switching tasks.
It can be observed that all comparison methods exhibit varying degrees of clothing inconsistency.
In the background maintaining task, these methods can only generate similar scenes but fail to ensure complete consistency.
While in the background switching task, these methods focus solely on the foreground subject, leading to deviation from the unique action prompts, \eg, the ``\textit{phone}'', ``\textit{costume character}'' in \cref{fig:quali} and the ``\textit{taking photos}'' in \cref{fig:compare} are missing. 
Additionally, we show the points and masks results in \cref{fig:point_mask}, to intuitively demonstrate the effectiveness of our point-matching and mask extraction methods.
More qualitative results are shown in the supplementary materials.

\myparagraph{Quantitative Results}
As shown in \cref{tab:quan}, compared to the consistent generation methods, across two evaluation tasks, our \textit{CharaConsist} achieves superior performance on most consistency metrics, especially in CLIP-I-bg and ID Sim. These results strongly validate the superiority of our method in maintaining character and scene consistency.
As the only exception, with the varying environments in the background switching task, the lower full-image CLIP-I score of our method is reasonable.
Compared to the identity-reference methods, \textit{CharaConsist} offers significant advantages in CLIP-I-fg and CLIP-I-bg scores, indicating our method's ability in preserving the details in character clothing and background environments.
Although our method does not surpass these methods on identity similarity, which is the training target for these methods, this is not contradictory.
As a training-free method, \textit{CharaConsist} can complement the identity-reference generation by addressing their shortcomings in clothing and environment consistency.
We will further discuss this in \cref{sec:discuss}.

Additionally, one concern is our method has lower CLIP-T score compared to other methods. This could be largely due to the domain differences stemming from the base model, as FLUX.1’s original results also show lower CLIP-T, which is further detailed in the ablation study and \cref{tab:incre}.

\myparagraph{Ablation Study}
Since this work is built on a different base model compared to the baseline methods, we conducted ablation study to analyze the impact of the base model itself and our proposed method.
For our \textit{CharaConsist} and the comparison methods, we evaluate both the original results of the corresponding base models and the increments on consistency.
The base models are SDXL, RealVisXL4.0, and FLUX.1 for ConsiStory, StoryDiffusion, and our \textit{CharaConsist}, respectively.
The results in \cref{tab:incre} demonstrate that, \textbf{first}, compared to the original results of FLUX.1, our method significantly enhances consistency, indicating its effectiveness.
\textbf{Second}, our method achieves the largest increment in consistency. This indicates that the superior consistency results shown in \cref{tab:quan} are not due to the new base model.
Instead, although FLUX.1 provides a notable improvement in image quality, its original results exhibit lower consistency compared to RealVisXL4.0 in almost all metrics.
\textbf{Furthermore}, The FLUX.1 exhibits a considerably lower CLIP-T score than other two base models, which may due to domain gaps in visual style.

\begin{table}[t]
\small
\centering
  \begin{tabular}{lcccccc}
    \Xhline{0.8pt}
    \multirow{2}{*}{Metrics} & \multicolumn{3}{c}{Base Model} & \multicolumn{3}{c}{Increment} \\
     & CsSt & StDiff & Ours & CsSt & StDiff & Ours \\
    \Xhline{0.5pt}
    CLIP-T & .368 & \textbf{.377} & .358 & \textbf{.0} & -.007 & -.002\\
    CLIP-I & .867 & \textbf{.882} & .874 & \textbf{.036} & .020 & \textbf{.036}\\
    CLIP-I-fg & .837 & \textbf{.855} & .842 & .039 & .027 & \textbf{.041}\\
    CLIP-I-bg & .868 & .877 & \textbf{.883} & .027 & .018 & \textbf{.033}\\
    ID Sim & .448 & \textbf{.546} & .476 & .130 & .096 & \textbf{.171}\\
    \Xhline{0.8pt}
  \end{tabular}
  \vspace{-3mm}
  \caption{\textbf{Consistency increments comparison}. The ``CsSt'' and ``StDiff'' refer to ConsiStory and StoryDiffusion respectively}
  \label{tab:incre}
  \vspace{-4mm}
\end{table}

\section{Conclusion and Discussion}
We propose \textit{CharaConsist} to enhance consistent text-to-image generation. By addressing the locality bias with point-tracking attention and adaptive token merge, it ensures fine-grained consistency under character action variations.
It also enables controllable background preserving or switching, broadening real-world applicability.

\label{sec:discuss}
\myparagraph{Limitation}
As a method focused on training-free consistent generation, our \textit{CharaConsist} can produce consistent contents aligned with textual prompts, but cannot take input identity as reference.
An ideal solution would be combining the trained identity-reference model with our method, in which the former could provide the ability to accept input identities and maintain higher facial similarity, while our method could compensate for their limitations in consistent background environments and character clothing.
This is an important direction for our future exploration.

\footnotesize{\paragraph{Acknowledgement.}~This project was supported by the National Natural Science Foundation of China (NSFC) under Grant No. 62472104.}

{
    \small
    \bibliographystyle{ieeenat_fullname}
    \bibliography{main}
}

\clearpage
\appendix
\normalsize

\twocolumn[\begin{@twocolumnfalse}
\begin{center}
  {\LARGE Appendix \par}
  \vspace{1em}
\end{center}
\vspace{2em}
\end{@twocolumnfalse}]

\section{Implementation Details}
Our \textit{CharaConsist} is built on the FLUX.1-dev \cite{flux} and maintains the original model configuration, which employs a standard Euler Sampler as in \cite{rf} with 50 sampling steps.

All of our additional modules are applied only to the single blocks of FLUX.1, which could provide sufficient consistency, without requiring manipulating all blocks. In contrast, manipulating all blocks could degrade the quality of the output images, occasionally leading to artifacts such as ghosting in some examples.
This conclusion is intuitive. In FLUX.1, the double blocks and single blocks, located in the early and later stages of the model respectively, resemble the structure of an encoder and decoder. The former encodes the input noisy latent, while the latter decodes it into the output. Therefore, we should manipulate the decoder to alter the output results.

\myparagraph{Point Matching and Mask Extraction}
The point matching relationships are obtained by calculating the cosine similarities between the intermediate-layer attention outputs from the frame image and those from the identity image. The similarities are further averaged across different layers to suppress fluctuations and achieve more stable and accurate outputs.
Similarly, our mask extraction strategy leverages the attention weights averaged across layers, deriving results by comparing the attention weight differences of image tokens to foreground and background text tokens.

\begin{figure}[t]
  \centering
  \includegraphics[width=1\linewidth]{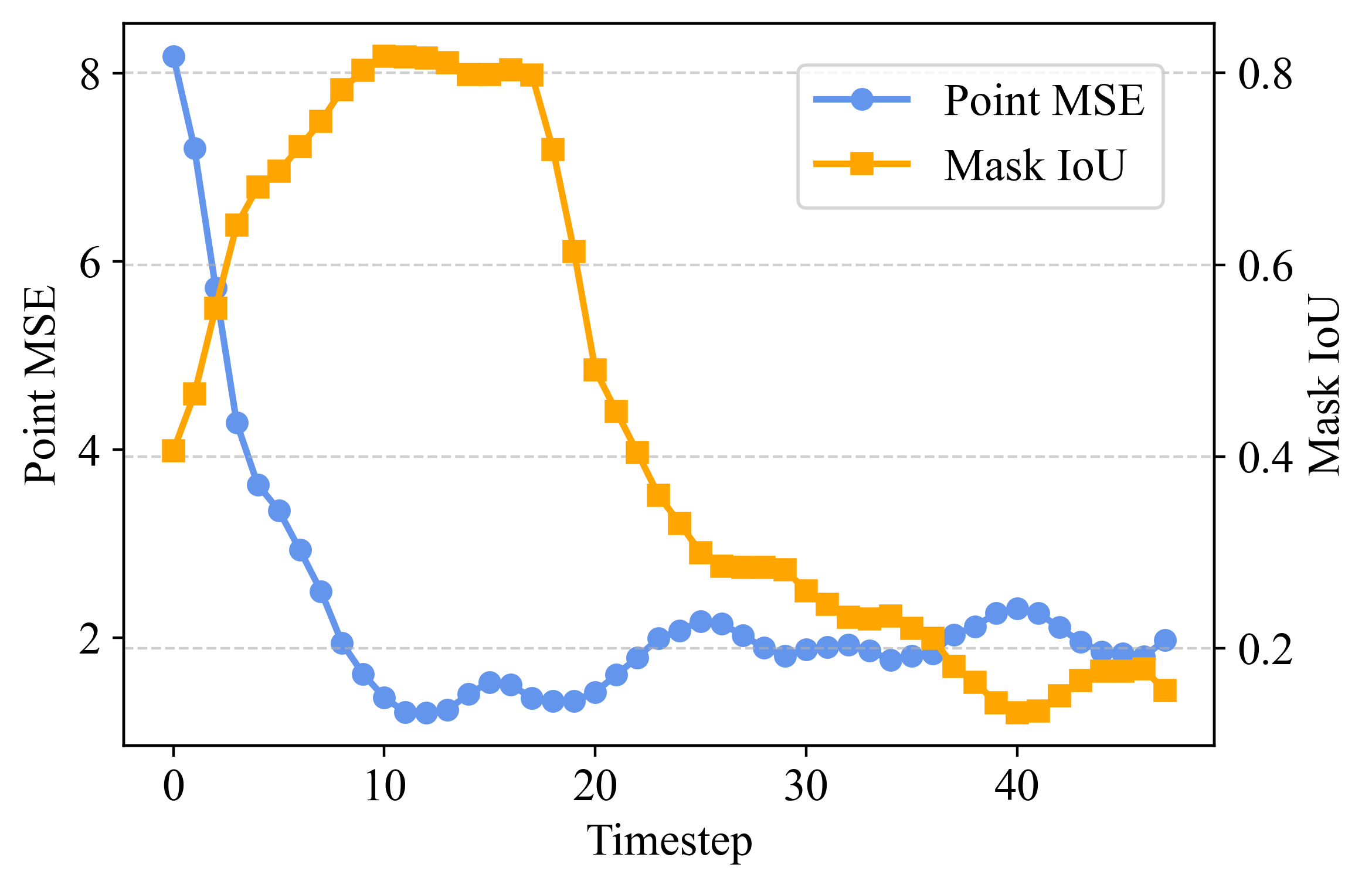}
  \caption{The accuracy of point matching and mask extraction across different timesteps of the generation process.}
  \label{fig:supp-point_mask_quan}
\end{figure}

However, it remains uncertain which timestep in the sampling process should be used to extract these results. As shown in \cref{fig:supp-point_mask_quan}, we manually annotated the point matching results and foreground masks for a set of generated outputs and made comparison with the automatically extracted results at different timesteps.
We calculated the mean squared error (MSE) for point matching and the intersection over union (IoU) for mask extraction. It can be observed that both metrics exhibit similar trends across timesteps, with a stable and accurate plateau appearing between steps 10 and 20. Consequently, we ultimately selected the point matching and mask results of the 11th timestep.

Additionally, in the mask extraction, the above quantitative results were based on the raw extracted masks, which includes small isolated foreground points and holes. To address this, we further applied morphological operations on the raw extracted masks: a $3\times3$ kernel erosion followed by a $5\times5$ kernel dilation.
The larger dilation parameter prevented the loss of foreground edge information.

\begin{figure*}
  \centering
  \includegraphics[width=1\linewidth]{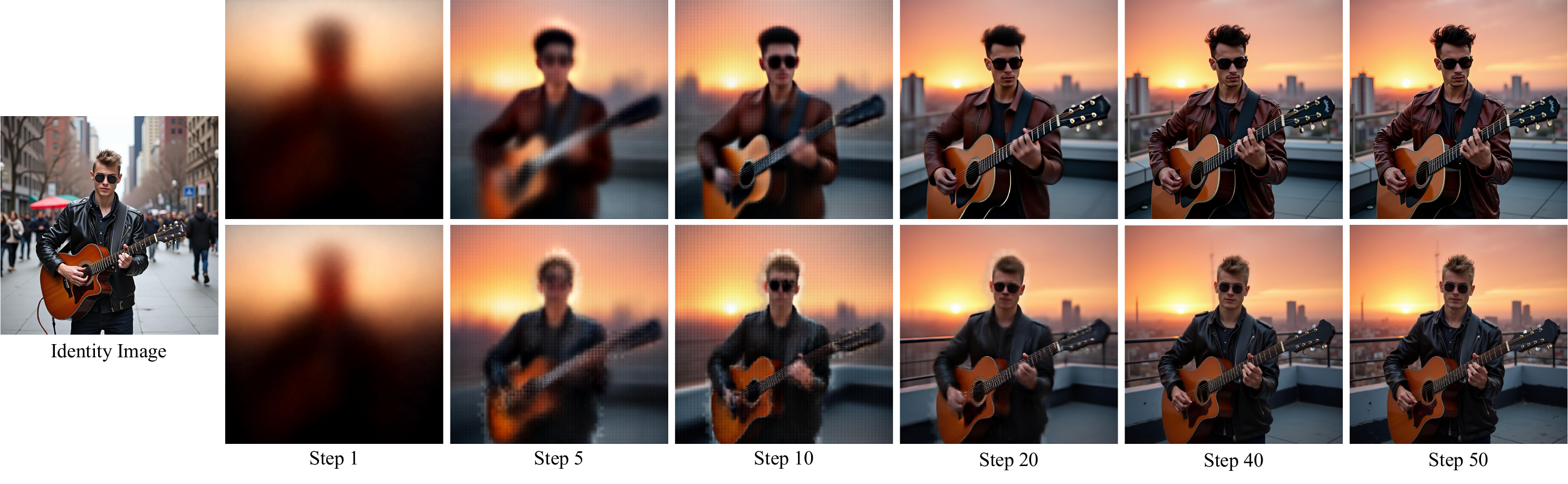}
  \caption{\textbf{Visualizations of generated images in intermediate timesteps.} Compared to the independently generated image, our \textit{CharaConsist} can progressively align the frame image with the identity image, especially in the  early timesteps.}
  \label{fig:supp-interm}
\end{figure*}

\myparagraph{Point-tracking Attention and Adaptive Token Merge}
As shown in \cref{fig:supp-interm}, the key features determining identity and details consistency, such as the character facial structure and hairstyle, are modeled at relatively early steps, while the later steps focus primarily on optimizing low-level details.
Therefore, we introduce the point-tracking attention and adaptive token merge starting from the 1st step of the sampling process.
These two modules are employed until the 40th step, as the last 10 steps have no significant impact on consistency, focusing only on refining low-level textures.

While as mentioned earlier, accurate point matching and masks results are obtained at the 11th step. Therefore, in the frame image generation process, we first perform the original sampling without additional modules from the 1st step to the 11th step and extract the corresponding point matching and mask results.
Then, based on these results, we introduce the proposed point-tracking attention and adaptive token merge, and perform the complete sampling process to achieve consistency.

\section{Additional Quantitative Results}
We provide additional ablation studies for each proposed component in \cref{tab:ablation}, with the identifiers refer to: (\textcolor{cyan}{a}) Point-tracking attention, (\textcolor{cyan}{b}) Adaptive token merge, and (\textcolor{cyan}{c}) Foreground and background masking.
Results show that (\textcolor{cyan}{a}) and (\textcolor{cyan}{b}) consistently enhance generation consistency, while performing masking in (\textcolor{cyan}{c}) can further improve background consistency. Moreover, our proposed modules do not significantly degrade the generation quality.

We also provide a user study in \cref{tab:user_study}. We invited 16 participants, each of whom was asked to evaluate 70 sets of images. For each set, they provided assessments along the following 5 dimensions: the consistency of facial ID, character clothing, and background, as well as the alignment of generated images with textual prompts and the aesthetics of images. The results show a clear preference for our method.

\begin{table}[t]
        \centering
        \setlength{\tabcolsep}{7.5pt}
        \begin{tabular}{lcccc}
            \Xhline{0.8pt}
            Metrics & T2I; & + (\textcolor{cyan}{a}); & + (\textcolor{cyan}{a}, \textcolor{cyan}{b}); & + (\textcolor{cyan}{a}, \textcolor{cyan}{b}, \textcolor{cyan}{c}) \\
            \Xhline{0.5pt}
            CLIP-I & 0.874 & 0.889 & 0.897 & 0.910 \\
            $\text{CLIP-I}_{fg}$ & 0.842 & 0.864 & 0.883 & 0.883 \\
            $\text{CLIP-I}_{bg}$ & 0.883 & 0.896 & 0.899 & 0.916 \\
            ID-Sim & 0.476 & 0.572 & 0.645 & 0.647 \\
            \Xhline{0.5pt}
            IAS & 0.839 & 0.835 & 0.828 & 0.831 \\
            IQS & 0.989 & 0.987 & 0.982 & 0.985 \\
            \Xhline{0.8pt}
        \end{tabular}
        \caption{Ablation studies.}
        \label{tab:ablation}
\end{table}

\begin{table}[t]
    \centering
    \begin{tabular}{lccc}
        \Xhline{0.8pt}
        Attributes & StoryDiffusion & ConsiStory & \textbf{\textit{Ours}} \\
        \Xhline{0.5pt}
        Facial ID & 26\% & 12\% & \textbf{62\%} \\
        Clothing & 21\% & 15\% & \textbf{64\%} \\
        Background & 2\% & 0\% & \textbf{98\%} \\
        \Xhline{0.5pt}
        Textual Align & 19\% & 24\% & \textbf{57\%} \\
        Aesthetics & 8\% & 9\% & \textbf{73\%}\\
        \Xhline{0.8pt}
    \end{tabular}
    \caption{User preference percent.}
    \label{tab:user_study}
\end{table}

\section{Additional Qualitative Results}
\begin{figure*}
  \centering
  \includegraphics[width=0.8\linewidth]{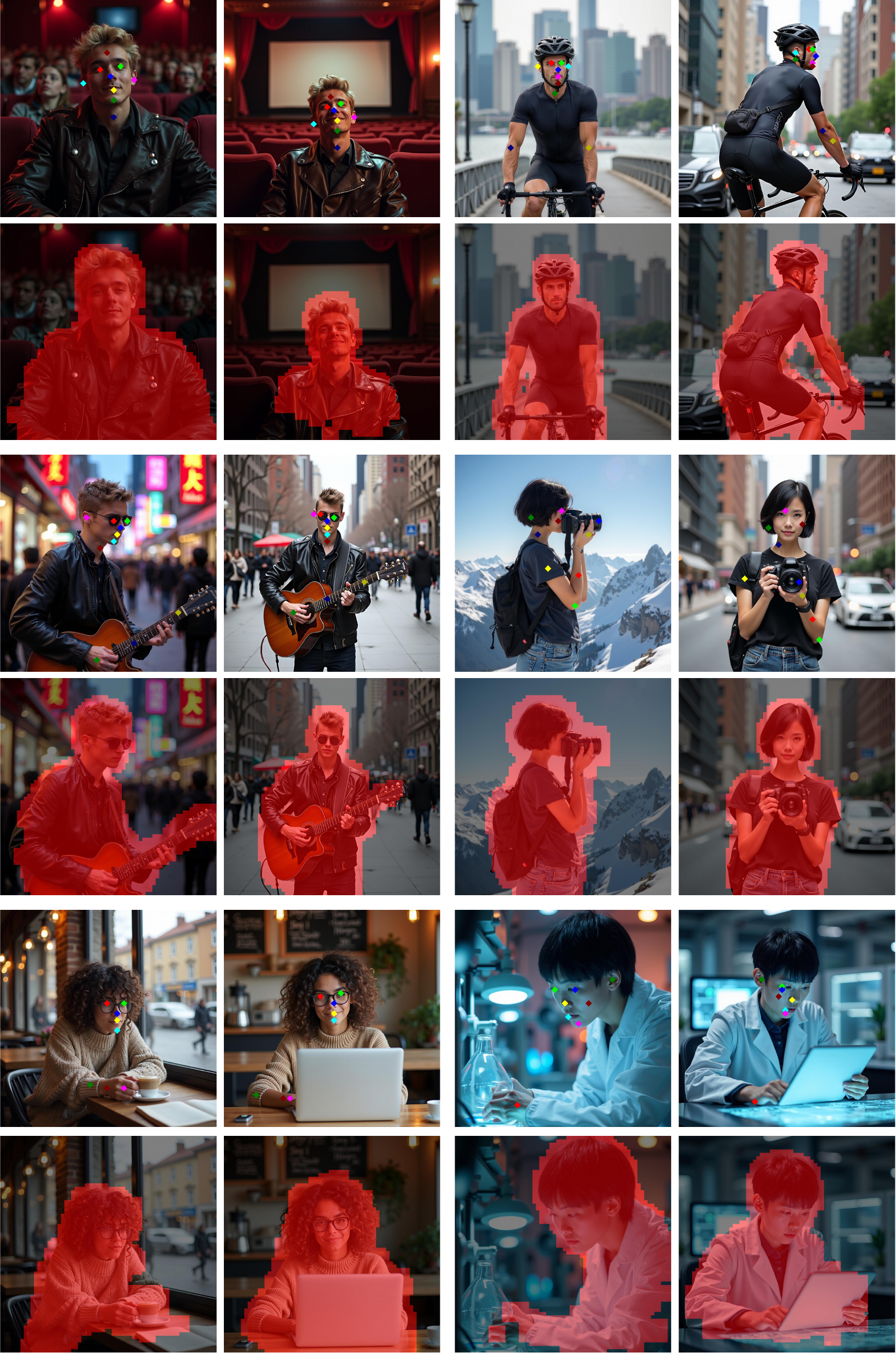}
  \caption{\textbf{Qualitative results of point matching and mask extraction.} The points in the first image of every pair are manually selected, while those in the second image are determined automatically by the matching relationship.}
  \label{fig:supp-point_mask_vis}
\end{figure*}

\begin{figure*}
  \centering
  \includegraphics[width=1.0\linewidth]{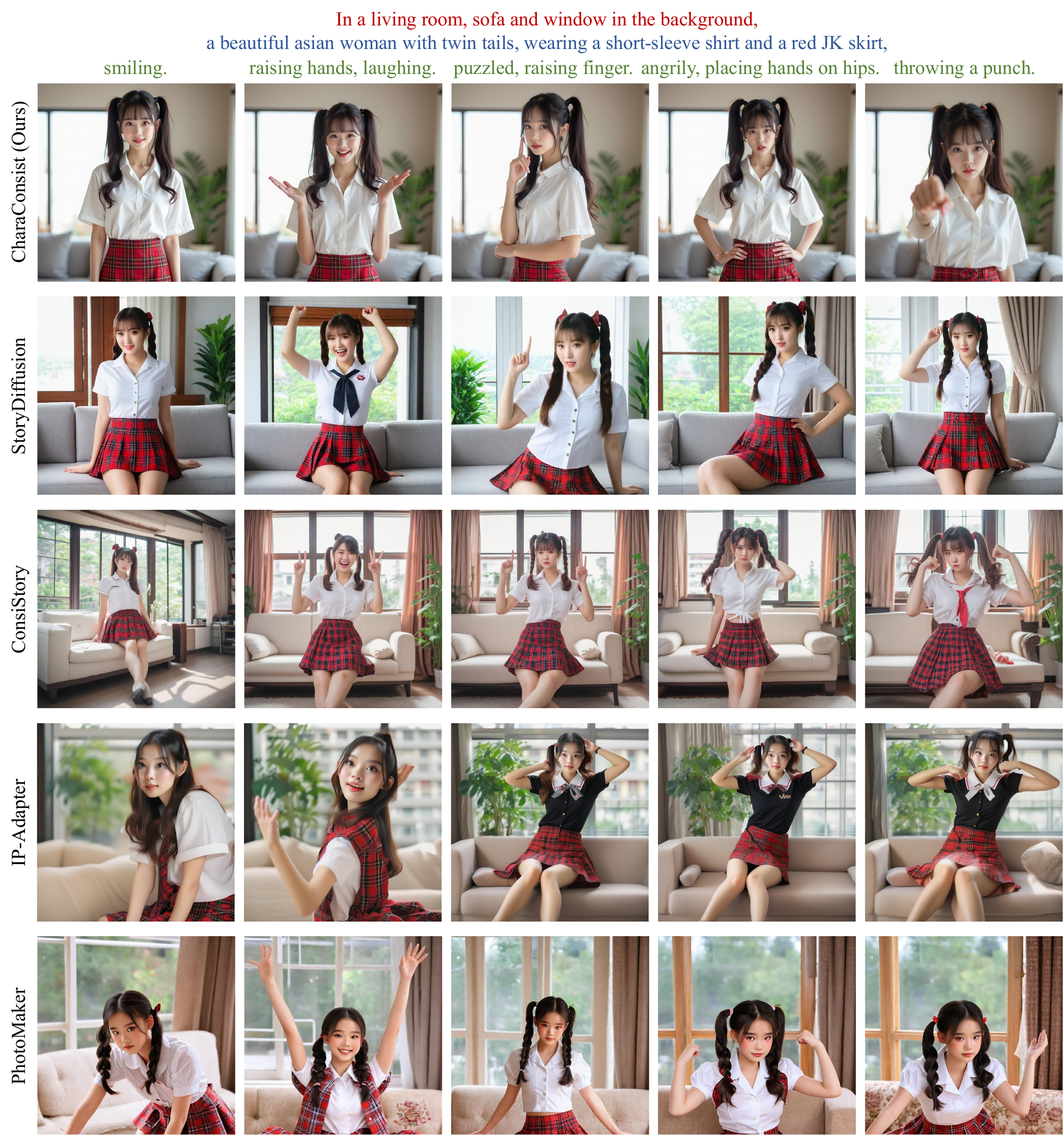}
  \caption{\textbf{Qualitative comparisons in the background maintaining task.} Comparison methods fail to maintain consistency in the environment and clothing. Furthermore, character's actions and expressions are overly similar across images, deviating from the unique prompt of each image. In contrast, our \textit{CharaConsist} achieves fine-grained consistency while preserving the flexibility of character actions.}
  \label{fig:supp-bg_fg_1}
\end{figure*}

\begin{figure*}
  \centering
  \includegraphics[width=1.0\linewidth]{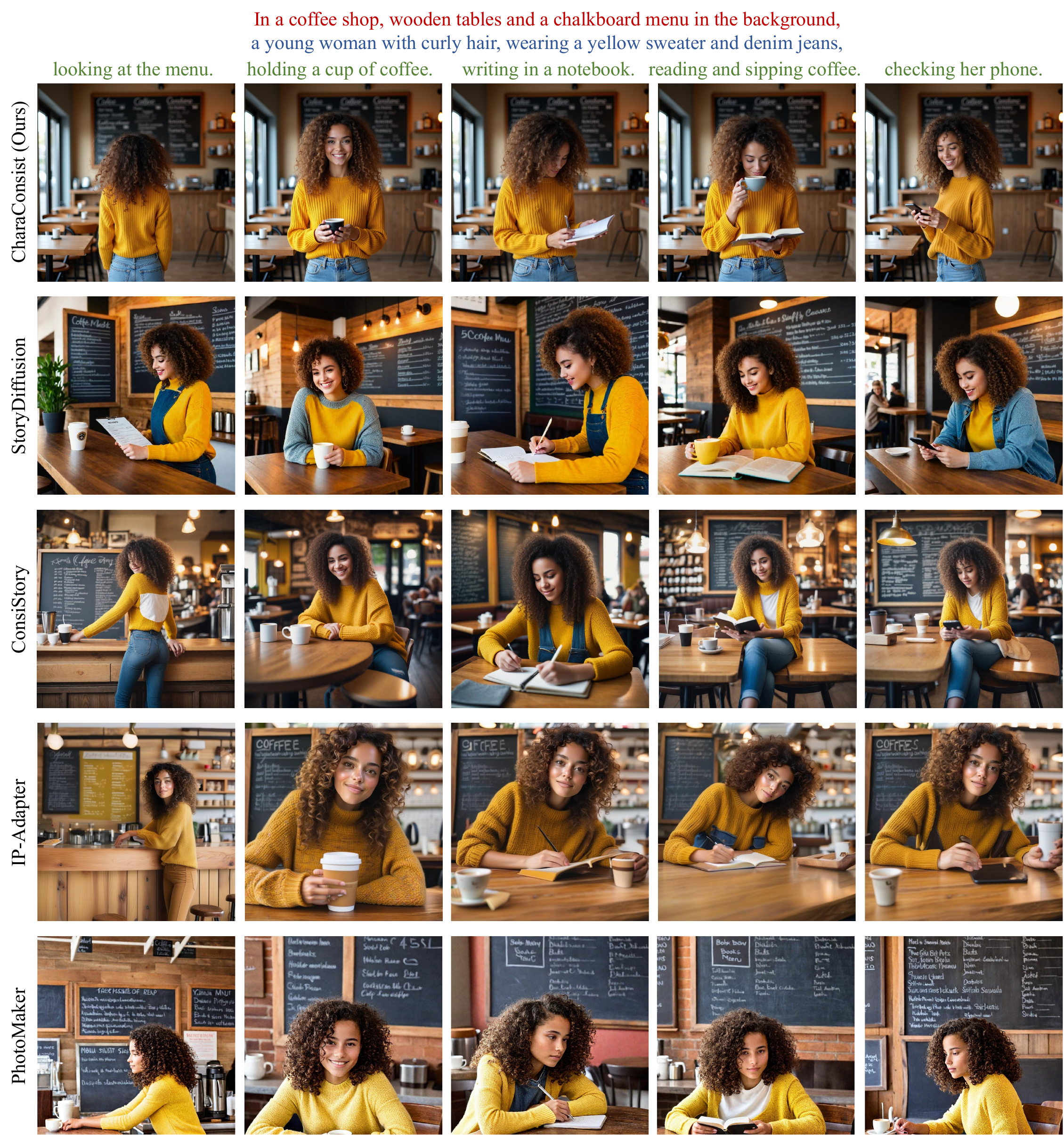}
  \caption{\textbf{Qualitative comparisons in the background maintaining task.} Our \textit{CharaConsist} achieves fine-grained consistency in the character identity, clothing, and background environments. Even the subtle details, for example, the writing on the chalkboard in the background, are well preserved by our method.}
  \label{fig:supp-bg_fg_2}
\end{figure*}

\begin{figure*}
  \centering
  \includegraphics[width=1.0\linewidth]{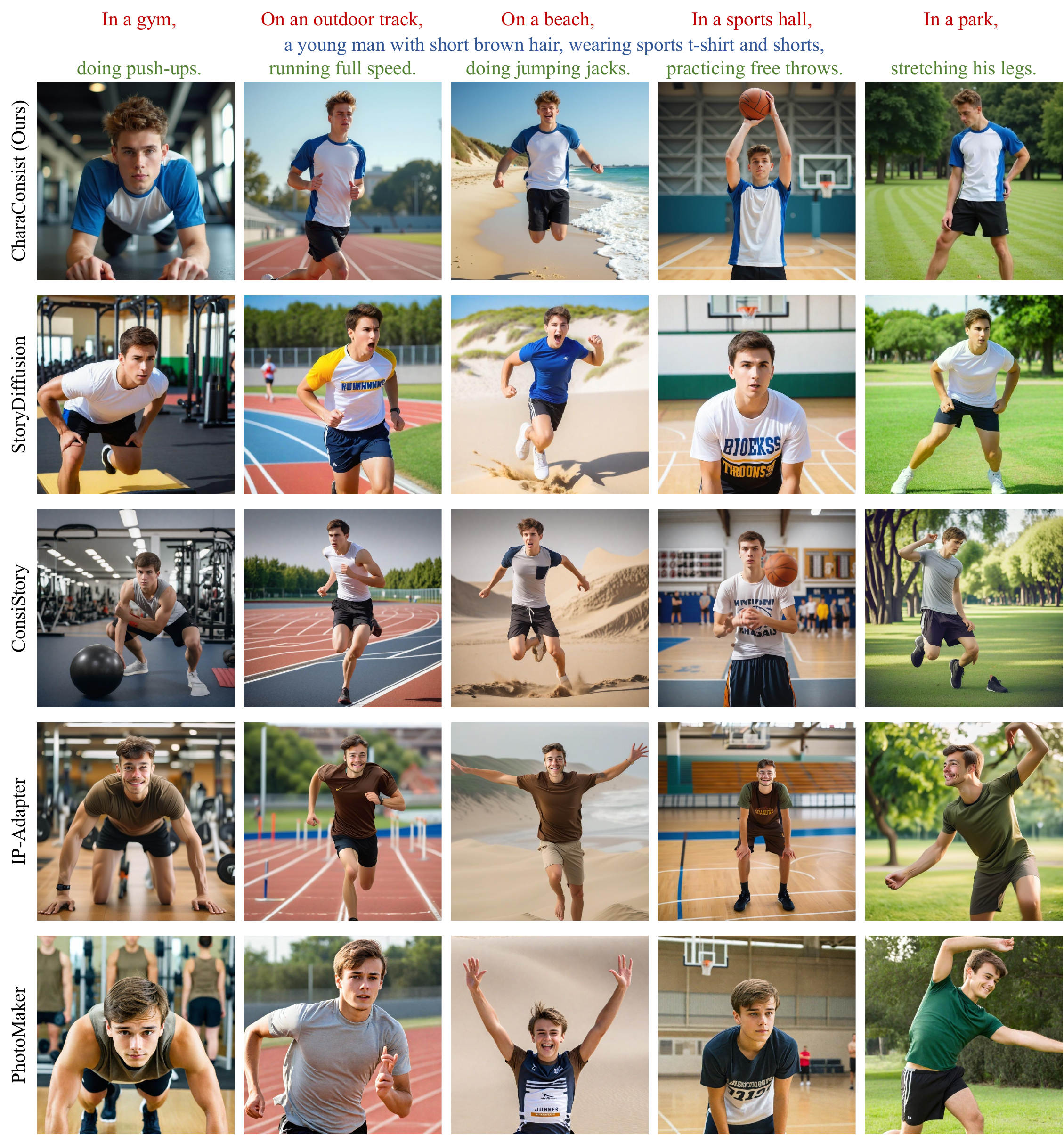}
  \caption{\textbf{Qualitative comparisons in the background switching task.} Our \textit{CharaConsist} can generate diverse backgrounds and character actions while maintaining consistency in the character's identity and clothing. While the comparison methods exhibit noticeable inconsistencies in character clothing.}
  \label{fig:supp-fg_1}
\end{figure*}

\begin{figure*}
  \centering
  \includegraphics[width=1.0\linewidth]{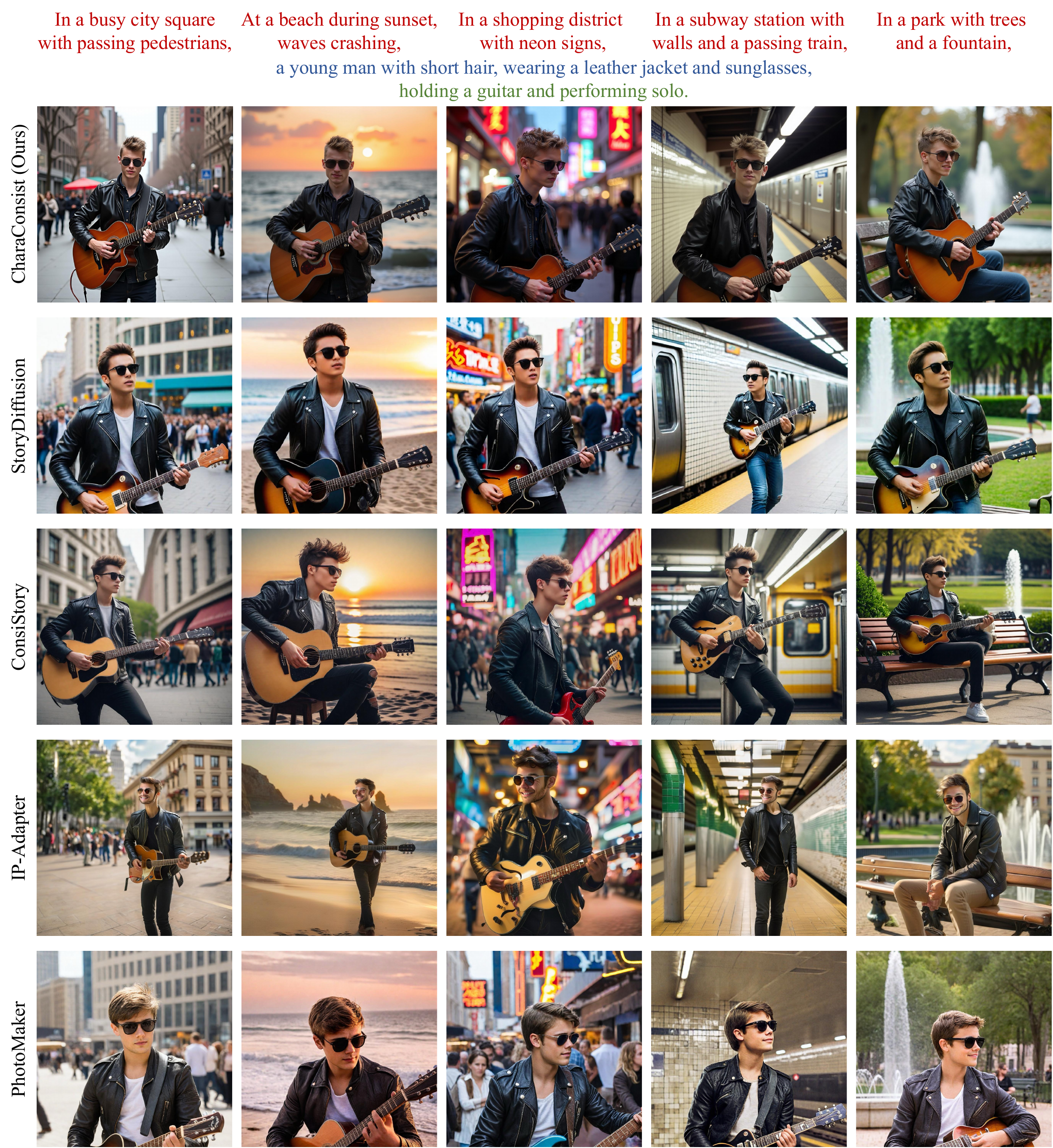}
  \caption{\textbf{Qualitative comparisons in the background switching task.} In different scenarios, our \textit{CharaConsist} effectively maintains consistency in the character's identity, clothing and the guitar. In contrast, the comparison methods show significant inconsistencies in these aspects, with the guitar element even missing in some examples.}
  \label{fig:supp-fg_2}
\end{figure*}

In the following pages, we provide additional visualizations and qualitative comparison results. In \cref{fig:supp-point_mask_vis}, we present more points and masks results to intuitively demonstrate the effectiveness of our point matching and mask extraction strategies. In \cref{fig:supp-bg_fg_1}, \cref{fig:supp-bg_fg_2}, \cref{fig:supp-fg_1}, and \cref{fig:supp-fg_2}, we showcase the qualitative comparison between our \textit{CharaConsist} and StoryDiffusion~\cite{storydiffusion}, ConsiStory~\cite{consistory}, IP-Adapter~\cite{ip-adapter}, and PhotoMaker~\cite{photo-maker}, in both the background maintaining and the background switching tasks. These results further demonstrate that our \textit{CharaConsist}'s superior performance in maintaining consistency in the various details of the foreground character identities, clothing, as well as the background environments.

\end{document}